# Unsupervised Segmentation of Overlapping Cervical Cell Cytoplasm

S L Happy, Swarnadip Chatterjee, and Debdoot Sheet, *Member, IEEE*

*Abstract*— Overlapping of cervical cells and poor contrast of cell cytoplasm are the major issues in accurate detection and segmentation of cervical cells. An unsupervised cell segmentation approach is presented here. Cell clump segmentation was carried out using the extended depth of field (EDF) image created from the images of different focal planes. A modified Otsu method with prior class weights is proposed for accurate segmentation of nuclei from the cell clumps. The cell cytoplasm was further segmented from cell clump depending upon the number of nucleus detected in that cell clump. Level set model was used for cytoplasm segmentation.

## I. Introduction

The Pap smear or Pap test is a procedure to detect cervical cancer. In Pap test, the cervical cells are examined under a microscope. For automatic analysis of Pap smear, the cervical cells need to be detected and segmented accurately. However, the presence of blood, inflammatory cells, mucus and other debris makes the detection process unreliable. Moreover, the cytoplasm of cervical cells have poor contrast and irregular shape. Segmenting the overlapping cervical cells is another major issue. An unsupervised method of cell segmentation is proposed here for accurate and automatic segmentation of overlapping cervical cells.

## II. Methods

The dataset provided in "The Second Overlapping Cervical Cytology Image Segmentation Challenge" contains of multi-layer cytology volume consisting of multi-focal images acquired from the same specimen. A stack of 20 images, each of size 1024x1024 pixels, at different focal planes are provided for each specimen image and they contain solitary or overlapping cervical cells with different degrees of overlap. The contrast and the texture of the cells are also not consistent.

Lu et al. [1] proposed an overlapping cell segmentation method in which cell clump segmentation is carried out first followed by nuclei detection. They used one nucleus per cell assumption to further segment the overlapping cells by

S L Happy and Debdoot Sheet are with the Electrical Engineering Department, Indian Institute of Technology Kharagpur, India (e-mail: happy@iitkgp.ac.in, debdoot@smst.iitkgp.ernet.in).

Swarnadip Chatterjee is with the Advanced Technology Development Center, Indian Institute of Technology Kharagpur, India. (e-mail: swarnadipchatterjeesmst@gmail.com).

minimizing an energy function. Similar method is adopted here. One extended depth of field (EDF) image [2] was first created out of each specimen volumes to obtain the number of cells present in the image as well as to obtain a tentative boundary of the cytoplasm. Further pre-processing on EDF image was carried out for cell clump segmentation followed by cytoplasm segmentation.

The proposed unsupervised approach implements three stages to successfully segment the overlapping cervical cell cytoplasm. First, the cell clumps were segmented from the EDF image. Second, each cell clump was processed individually to segment the nucleus and obtain the nucleus count in that cell clump. Third, depending upon the nucleus position and nucleus count, distance regularized level set model [3] was used in each cell clumps for cytoplasm segmentation. The number of cells in a specimen was assumed to be equal to the count of nuclei, thereby segmenting the cell cytoplasm from the previously segmented cell clump.

### A. Cell clump segmentation

The EDF image was preprocessed to remove noise and unwanted small grain like structures. First, a median filter of size $5 \times 5$ was applied followed by adaptive histogram equalization. Further, H-maxima transform was applied and binary image was obtained using the regional maxima of image. The unwanted small areas in binary image were removed by removing all connected components having less than a threshold which was approximately equal to the area of smallest cytoplasm present in an image. Each connected component was considered as a cell clump.

### B. Modified Otsu method with class prior probability

Nucleus segmentation is a challenging task since the contrast of the overlapping cytoplasm is not uniform. We propose a modified Otsu threshold approach for nucleus segmentation. Otsu's method assumes bimodal distribution of gray-level intensities and the threshold can be obtained by minimizing the within-class variances [4], given by

$$T = \arg \min_{1 \leq T < L_{max}} \{\sigma_w^2(T)\}$$

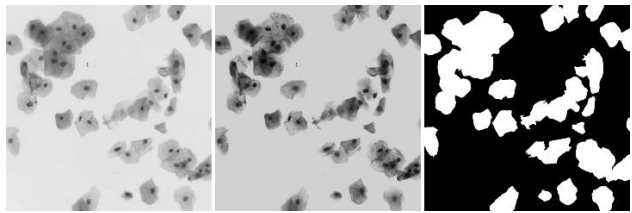

Fig. 1. Results of scene segmentation. (a) The EDF image, (b) prepossessed image for segmentation, (c) cell clump segmentation.

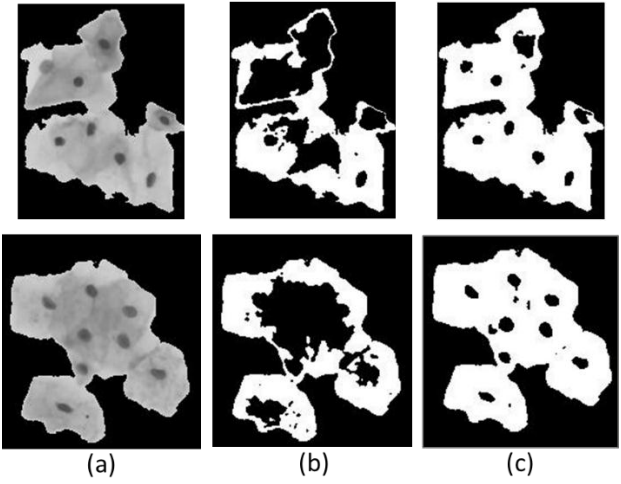

Fig. 2. Nucleus segmentation, (a) cell clump, (b) Otsu's method, (c) modified Otsu with prior class probabilities (here $\alpha = 0.05$).

The within-class variance ($\sigma_w^2$) can be written in terms of class variance ($\sigma_i$) and class probabilities ($\omega_i$) as

$$\sigma_w^2(t) = \omega_1(t)\sigma_1^2(t) + \omega_2(t)\sigma_2^2(t)$$

Here, $T$ is the threshold separating the classes. Given the histogram ($h$) of an image, we can find the probability of each pixel $p(i) = \frac{h(i)}{Total\ Pixels}, i = 1, \ldots, L_{max}$ ($L_{max}$ = maximum pixel label). The class probabilities can be calculated from the image histogram as

$$\omega_1(t) = \sum_{i=1}^{t} p(i) \text{ and } \omega_2(t) = \sum_{i=t+1}^{L_{max}} p(i)$$

Xu et al. [4] reported that Otsu threshold deviates from the intersection point toward the center of the class having higher variance. Hence, it is applicable for segmentation if both classes have approximately equal within-class variances. However, nucleus segmentation from the cell clump falls under the category of unequal class probabilities. Therefore, we modify the distribution of the pixels based on the prior class probabilities to segment nucleus accurately.

We propose a prior probability based modified Otsu method to find suitable threshold when the class probability of one class is very less compared to the other. Given the prior class probability of the class containing dark pixels, $\alpha$, we determine the maximum value of $\ell$ such that $\sum_{i=1}^{\ell} p(i) < \alpha$. The pixel distributions are further modified as,

$$p'(i) = \begin{cases} p(i) * (1 - \alpha) & \text{if } i < \ell \\ p(i) * \alpha & \text{otherwise} \end{cases}$$

$$p_{new}(i) = \frac{p'(i)}{\sum_i p'(i)}$$

If $\alpha$ is small, then the pixel probabilities up to $\ell$ is given a higher weight than the rest and vice versa. The threshold is further determined using $p_{new}(i)$ as the pixel probabilities. Thus, the class distribution is modified to drag the threshold to the intersection point. Fig. 2 shows nucleus segmentation results using prior probability of nucleus region in a cell clump as 0.05.

*C. Cytoplasm segmentation*

The cell clump is further segmented using the distance regularized level set model [3] which propagates outward from the position of cell nucleus. If the number of nucleus in the connected region is one, then the connected region is considered as one cell. In case of multiple nucleus in a clump, we initialized the cell region as a disc surrounding the nucleus which further propagated toward the cytoplasm edges.

III. RESULTS

The proposed method was evaluated using the training set provided in ISBI 2015 challenge. The training set consisted of depth images of 8 specimens with ground truth of cell cytoplasm. The cytoplasm segmentation results obtained by the proposed method on the training set are provided in Table I.

TABLE I. QUANTITATIVE RESULTS ON THE TRAINING SETS

| Dataset | DC | TPR | FPR |
|---|---|---|---|
| Training set | 0.8607 ± 0.0733 | 0.8827 ± 0.0944 | 0.0014 ± 0.0014 |

(DC: Dice Coefficient; TPR: True positive rate (pixel-level); FPR: False positive rate (pixel-level))

Dice coefficient of 0.86 is achieved with our algorithm as shown in Table I with TPR 0.88. The proposed method segments both nucleus and cytoplasm accurately.